\documentclass[10pt,twocolumn,letterpaper]{article}

\usepackage[pagenumbers]{cvpr} 

\usepackage{arydshln}
\usepackage[dvipsnames]{xcolor}

\definecolor{cvprblue}{rgb}{0.21,0.49,0.74}
\usepackage[pagebackref,breaklinks,colorlinks,citecolor=cvprblue]{hyperref}
\usepackage{multirow}
\usepackage{bbm}
\usepackage{animate}
\usepackage{listings}
\usepackage{pifont}

\newcommand{\cmark}{\ding{51}}%
\title{Towards A Better Metric for Text-to-Video Generation}

\author{
Jay Zhangjie Wu$^{1\ast}$\quad Guian Fang$^{1\ast}$\quad Haoning Wu$^{4\ast}$
\quad Xintao Wang$^{3}$\quad Yixiao Ge$^{3}$
\\ 
Xiaodong Cun$^{3}$\quad David Junhao Zhang$^{1}$\quad Jia-Wei Liu$^{1}$\quad Yuchao Gu$^{1}$\quad Rui Zhao$^{1}$
\\
Weisi Lin$^{4}$\quad Wynne Hsu$^{2}$\quad Ying Shan$^{3}$\quad Mike Zheng Shou$^{1}$
\\ 
\vspace{-0.6em} \\ 
$^1$Show Lab, $^2$National University of Singapore\quad $^3$ARC Lab, Tencent PCG \\ 
$^4$Nanyang Technological University 
\\
\vspace{-0.6em} \\ 
{\url{https://showlab.github.io/T2VScore}}
}

\begin{document}
\maketitle

{
  \renewcommand{\thefootnote}%
    {\fnsymbol{footnote}}
  \footnotetext[0]{$^\ast$Equal contribution.}
}

\begin{abstract}
Generative models have demonstrated remarkable capability in synthesizing high-quality text, images, and videos. For video generation, contemporary text-to-video models exhibit impressive capabilities, crafting visually stunning videos. Nonetheless, evaluating such videos poses significant challenges. Current research predominantly employs automated metrics such as FVD, IS, and CLIP Score. However, these metrics provide an incomplete analysis, particularly in the temporal assessment of video content, thus rendering them unreliable indicators of true video quality. Furthermore, while user studies have the potential to reflect human perception accurately, they are hampered by their time-intensive and laborious nature, with outcomes that are often tainted by subjective bias. In this paper, we investigate the limitations inherent in existing metrics and introduce a novel evaluation pipeline, the Text-to-Video Score (T2VScore). This metric integrates two pivotal criteria: (1) Text-Video Alignment, which scrutinizes the fidelity of the video in representing the given text description, and (2) Video Quality, which evaluates the video's overall production caliber with a mixture of experts. Moreover, to evaluate the proposed metrics and facilitate future improvements on them, we present the \textbf{TVGE} dataset, collecting human judgements of 2,543 text-to-video generated videos on the two criteria.
Experiments on the TVGE dataset demonstrate the superiority of the proposed T2VScore on offering a better metric for text-to-video generation. The code and dataset will be open-sourced.
\end{abstract}    
\section{Introduction}
\label{sec:intro}

\begin{figure}[t!]
    \centering
    \animategraphics[width=\linewidth,loop]{8}{fig/teaser/}{00}{10}
    \caption{\textbf{\texttt{T2VScore}}: We measure text-conditioned generated videos from two essential perspectives: \textit{text alignment} and \textit{video quality}. Our proposed \texttt{T2VScore} achieves the highest correlation with human judgment.
    \emph{We encourage readers to \textcolor{magenta}{click and play} using Adobe Acrobat.}}
    \label{fig:teaser}
    \vspace{-1em}
\end{figure}

Text-to-video generation marks one of the most exciting achievements in generative AI, with awesome video generative models coming out from companies~\cite{pika, ho2022imagenvideo, singer2022make, videoLDM, gen2} and opensource community~\cite{show1, zeroscope, modelscope, video_crafter}. These models, equipped with the ability to learn from vast datasets of text-video pairs, can generate creative video content that can range from simple animations to complex, lifelike scenes. 

To assess text-conditioned generated videos, most existing studies employ objective metrics like Fréchet Video Distance (FVD)~\cite{fvd} and Video Inception Score (IS)~\cite{videoIS} for \textit{video quality}, and CLIPScore~\cite{clip} for \textit{text-video alignment}. However, these metrics have limitations. FVD and Video IS are unsuitable for open-domain video generation due to their Full-Reference nature. Meanwhile, the CLIP Score computes an average of per-frame text-image similarities using image CLIP models, overlooking important temporal motion changes in videos. This leads to a mismatch between these objective metrics and human perception, as evident in recent studies~\cite{otani2023toward, liu2023fetv}. Current studies also incorporate subjective user evaluations for text-to-video generation. However, conducting large-scale human evaluations is labor-intensive and, therefore, not practical for widespread, open comparisons. To address this, there is a need for fine-grained automatic metrics tailored for evaluating text-guided generated videos.

In this work, we take a significant step forward by introducing \texttt{T2VScore}, a novel automatic evaluator specifically designed for text-to-video generation. \texttt{T2VScore} assesses two essential aspects of text-guided generated videos: \textit{text-video alignment} (\ie, how well does the video match the text prompt?), and \textit{video quality} (\ie, how good is the quality of the synthesized video?). Two metrics are then introduced: 1) \texttt{T2VScore-A} evaluates the correctness of all spatial and temporal elements in the text prompt by querying the video using cutting-edge vision-language models; 2) \texttt{T2VScore-Q} is designed to predict a robust and generalizable quality score for text-guided generated videos via a combo of structural and training strategies.

To examine the reliability and robustness of the proposed metrics in the evaluation of text-guided generated videos, we present the \textbf{T}ext-to-\textbf{V}ideo \textbf{G}eneration \textbf{E}valuation (TVGE) dataset. This dataset gathers extensive human opinions on two key aspects: \textit{text-video alignment} and \textit{video quality}, as investigated in our \texttt{T2VScore}. The TVGE dataset will serve as an open benchmark for assessing the correlation between automatic metrics and human judgments. Moreover, it can help automatic metrics to better adapt to the domain of text-guided generated videos. Extensive experiments on the TGVE dataset demonstrate better alignment of our \texttt{T2VScore} with human judgment compared to all baseline metrics.

To summarize, we make the following contributions:

\begin{itemize}
    \setlength\itemsep{0.8em}
    \item We introduce \texttt{T2VScore} as a novel evaluator dedicated to automatically assessing text-conditioned generated videos, focusing on two key aspects: \textit{text-video alignment} and \textit{video quality}.
    \item We collect the Text-to-Video Generation Evaluation (TVGE) dataset, which is posited as the first open-source dataset dedicated to benchmarking and enhancing evaluation metrics for text-to-video generation.
    \item We validate the inconsistency between current objective metrics and human judgment on the TVGE dataset. Our proposed metrics, \texttt{T2VScore-A} and \texttt{T2VScore-Q}, demonstrate superior performance in correlation analysis with human evaluations, thereby serving as more effective metrics for evaluating text-conditioned generated videos.
\end{itemize}

\section{Related Work}
\label{sec:related_work}

\subsection{Text-to-Video Generation}
Diffusion-based models have been widely explored to achieve text-to-video generation~\cite{vdm, videofactory, pyoco, videoLDM, lvdm, show1, magicvideo, ho2022imagenvideo, tuneavideo, modelscope, videogen, zhao2023motiondirector, lavie, nuwaxl}.
VDM~\cite{vdm} pioneered the exploration of the diffusion model in the text-to-video generation, in which a 3D version of U-Net~\cite{unet} structure is explored to jointly learn the spatial and temporal generation knowledge. 
Make-A-Video~\cite{singer2022make} proposed to learn temporal knowledge with only unlabeled videos. Imagen Video~\cite{ho2022imagenvideo} built cascaded diffusion models to generate video and then spatially and temporally up-sample it in cascade. 
PYoCo~\cite{pyoco} introduced the progressive noise prior model to preserve the temporal correlation and achieved better performance in fine-tuning the pre-trained text-to-image models to text-to-video generation.
The subsequent works, LVDM~\cite{lvdm} et al., further explored training a 3D 
U-Net in latent space to reduce training complexity and computational costs. 
These works can be classified respectively as pixel-based models and latent-based models.
Show-1~\cite{show1} marks the first integration of pixel-based and latent-based models for video generation. It leverages pixel-based models for generating low-resolution videos and employs latent-based models to upscale them to high resolution, combining the advantages of high efficiency from latent-based models and superior content quality from pixel-based models. 
Recently, the text-to-video generation products, such as Gen-2~\cite{gen2}, Pika~\cite{pika}, and Floor33~\cite{floor33}, and the open-sourced foundational text-to-video diffusion models, such as ModelScopeT2V~\cite{wang2023modelscope}, ZeroScope~\cite{zeroscope}, VideoCrafter~\cite{video_crafter}, have democratized the video generation, garnering widespread interest from both the community and academia.

\subsection{Evaluation Metrics}


\paragraph{Image Metrics.} 
Image-level metrics are widely utilized to evaluate the frame quality of generated videos. These include Peak Signal-to-Noise Ratio (PSNR)~\cite{psnr}, and Structural Similarity Index (SSIM)~\cite{SSIM}, Learned Perceptual Image Patch Similarity (LPIPS)~\cite{zhang2018unreasonable}, Fréchet Inception Distance (FID)~\cite{fid}, and CLIP Score~\cite{clip}. Among them, the PSNR~\cite{psnr}, SSIM~\cite{SSIM}, and LPIPS~\cite{zhang2018unreasonable} are mainly employed to evaluate the quality of reconstructed video frames by comparing the difference between generated frames and original frames. Specifically, PSNR~\cite{psnr} is the ratio between the peak signal and the Mean Squared Error (MSE)~\cite{mse}. SSIM~\cite{SSIM} evaluates brightness, contrast, and structural features between generated and original images. LPIPS~\cite{zhang2018unreasonable} is a perceptual metric that computes the distance of image patches in the latent feature space. FID~\cite{fid} utlizes the InceptionV3~\cite{incpetionv3} to extract feature maps from normalized generated and real-world frames, and computes the mean and covariance matrices for FID~\cite{fid} scores. CLIP Score~\cite{clip} measures the similarity of the CLIP features extracted from the images and texts, and it has been widely employed in text-to-video generation or editing tasks~\cite{singer2022make, show1, videoLDM, tuneavideo, zhao2023motiondirector, liu2023dynvideo, gu2023videoswap}.

\vspace{-1em}
\paragraph{Video Metrics.} 
In contrast to the frame-wise metrics, video metrics focus more on the comprehensive evaluation of the video quality. Fréchet Video Distance (FVD)~\cite{fvd} utilizes the Inflated-3D Convnets (I3D)~\cite{i3d} pre-trained on Kinetics~\cite{kinetic} to extract the features from videos, and compute their means and covariance matrices for FVD scores. Differently, Kernel Video Distance (KVD)~\cite{kvd} computes the Maximum Mean Discrepancy (MMD)~\cite{MMD} of the video features extracted using I3D~\cite{i3d} to evaluate the video quality. Video Inception Score (Video IS)~\cite{videoIS} computes the inception score of videos with the features extracted from C3D~\cite{c3d}. Frame Consistency CLIP Score~\cite{clip} calculates the cosine similarity of the CLIP image embeddings for all pairs of video frames to measure the consistency of edited videos~\cite{tuneavideo, controlvideo, SimDA, wu2023cvpr, liu2023dynvideo, gu2023videoswap}.

\subsection{Video Quality Assessment}

\noindent State-of-the-arts on video quality assessment (VQA) have been predominated by learning-based approaches~\cite{wu2023temporal,vsfa,zhang2022bvqa}. Typically, these approaches leverage pre-trained deep neural networks as feature extractors and use human opinions as supervision to regress these features into quality scores. Some most recent works~\cite{wu2022fastvqa,wu2023dover,wu2022fastervqa} have adopted a new strategy that uses a large VQA database on natural videos~\cite{pvq} to learn better feature representations for VQA, and then transfer to diverse types of videos with only a few labeled videos available. This strategy has been validated as an effective way to improve the prediction accuracy and robustness on relatively small VQA datasets for enhanced videos~\cite{ntire2023vqa} and computer-generated contents~\cite{ytugc}. In our study, we extend this strategy for evaluating text-conditioned generated videos, bringing a more reliable and generalizable quality metric for text-to-video generation.

Despite leveraging from large video quality databases, several recent works~\cite{clipiqa,wu2023qbench,wu2023bvqi,bvqiplus} have also explored to adopt multi-modality foundation models \textit{e.g.} CLIP~\cite{clip} for VQA. With the text prompts as natural quality indicators (\textit{e.g. good/bad}), these text-prompted methods prove superior abilities on zero-shot or few-shot VQA settings, and robust generalization among distributions. Inspired by existing studies, the proposed quality metric in T2VScore also ensembles a text-prompted structure, which is proved to better align with human judgments on videos generated by novel generators that are not seen during training.

\subsection{QA-based Evaluation}

Recent studies have emerged around the idea of using Visual Question Answering (VQA) to test the accuracy of advanced AI models. TIFA~\cite{hu2023tifa} utilizes GPT-3~\cite{brown2020language} to create questions in various areas, such as color and shape, and checks the answers with VQA systems like mPLUG~\cite{li2022mplug}. $\mathrm{VQ^2A}$\cite{changpinyo2022vq2a} makes VQA more reliable by synthesizing new data and employing high-quality negative sampling. VPEval\cite{Cho2023VPT2I} improves this process through the use of object detection and Optical Character Recognition (OCR), combining these with ChatGPT for more controlled testing. However, these methods have not yet explored videos, where both spatial and temporal elements should be evaluated. We are adding specific designs to the temporal domain to improve VQA for video understanding. This provides a more comprehensive method for evaluating text-video alignment from both space and time.
\section{Proposed Metrics}\label{sec:metrics}

We introduce two metrics to evaluate text-guided generated videos, focusing on two essential dimensions: Text Alignment (\cref{sec:text_alignment}) and Video Quality (\cref{sec:video_quality}).

\subsection{Text Alignment}\label{sec:text_alignment}

\begin{figure}[t!]
    \centering
    \animategraphics[width=\linewidth,loop]{8}{fig/text_alignment/}{00}{16}
    \caption{\textbf{Pipeline for Calculating \texttt{T2VScore-A}:} We input the text pormpt into large language models (LLMs) to generate questions and answers. Utilizing CoTracker~\cite{cotracker}, we extract the auxiliary trajectory, which, along with the input video, is fed into multimodal LLM (MLLMs) for visual question answering (VQA). The final \texttt{T2VScore-A} is measured based on the accuracy of VQA.
    \emph{Please \textcolor{magenta}{click and play} using Adobe Acrobat.}}
    \label{fig:text_alignment}
    \vspace{-1em}
\end{figure}

State-of-the-art multimodal large language models (MLLMs) have demonstrated human-level capabilities in both visual and textual comprehension and generation. Here, we introduce a framework for assessing the text-video alignment using these MLLMs. An overview of our text alignment evaluation process is presented in \cref{fig:text_alignment}.

\vspace{-1em}
\paragraph{Entity Decomposition in Text Prompt.} Consider a text prompt denoted as $\mathcal{P}$. Our initial step involves parsing $\mathcal{P}$ into distinct semantic elements, represented as $e_i$. We then identify the hierarchical semantic relationships among these elements, forming entity tuples $\{(e_i, e_j)\}$. Here, $e_j$ is semantically dependent on $e_i$ to form a coherent meaning. For instance, the tuple $(\text{dog}, \text{a})$ implies that the article ``a" is associated with the noun ``dog", while $(\text{cat}, \text{playing soccer})$ suggests that the action ``playing soccer" is attributed to the ``cat". Elements that exert a global influence over the entire prompt, like \emph{style} or \emph{camera motion}, are categorized under a \texttt{global} element. This structuring not only clarifies the interconnections within the elements of a text prompt but also implicitly prioritizes them based on their hierarchical significance. For instance, mismatching an element that holds a higher dependency rank would result in a more substantial penalty on the final text alignment score.

\vspace{-1em}
\paragraph{Question/Answer Generation with LLMs.} Our main goal is to generate diverse questions that cover all elements of the text input evenly. Drawing inspiration from previous studies~\cite{hu2023tifa}, for a text prompt $\mathcal{P}$, we utilize large language models (LLMs) to generate question-choice-answer tuples $\{Q_i, C_i, A_i\}_{i=1}^N$, as depicted on the top of \cref{fig:text_alignment}. Different from prior work focusing on text-image alignment, we emphasize the temporal aspects, such as object trajectory and camera motion, which are unique and essential for evaluating text alignment in dynamic video contexts. We employ a single-pass inference using in-context learning with GPT-3.5~\cite{brown2020language, wei2022chain} to generate both questions and answers. We manually curate 3 examples and use them as in-context examples for GPT-3.5 to follow. The complete prompt used for generating question and answer pairs can be found in the supplementary.

\begin{figure}[t!]
    \centering
    \includegraphics[width=0.99\linewidth]{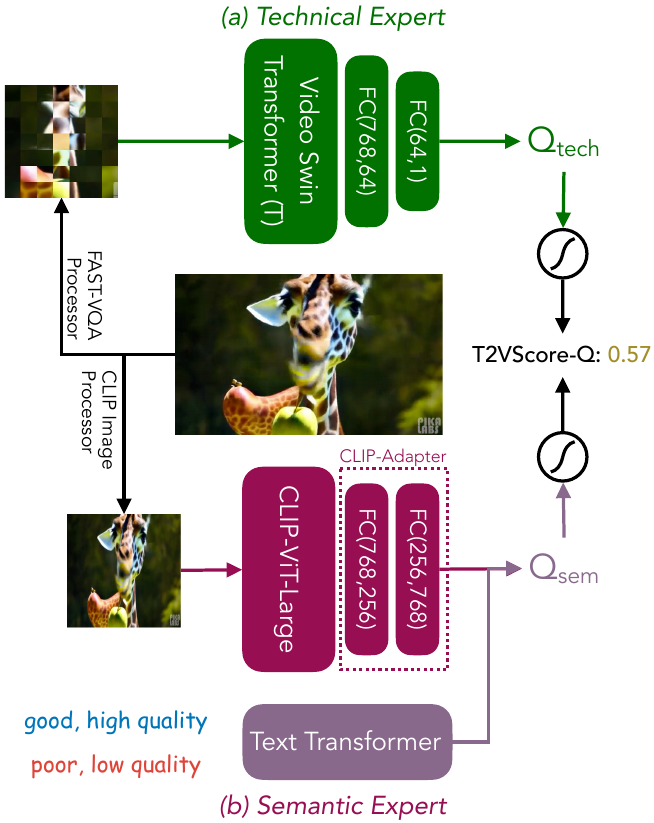}
    \caption{\textbf{Pipeline for Calculating \texttt{T2VScore-Q}:} a mixture of a \textbf{\textit{technical expert}} (a) to capture \textit{spatial} and \textit{temporal} technical distortions, and a text-prompted \textbf{\textit{semantic expert}} (b).}
    \label{fig:vqstructure}
    \vspace{-1em}
\end{figure}

\vspace{-1em}
\paragraph{Video Question Answering with Auxiliary Trajectory.} Most open-domain vision-language models are image-centric~\cite{blip2, yang2023dawn, minigpt4, llava, llava2, instructblip}, with only a few focusing on video~\cite{videochat, videochatgpt, mplugowl2}. These VideoLLMs often struggle with fine-grained temporal comprehension, as evidenced by their performance on benchmarks like SEED-Bench~\cite{seed-bench}. To address this, we introduce the use of auxiliary trajectories, generated by off-the-shelf point tracking models (e.g., CoTracker~\cite{cotracker} and OmniMotion~\cite{wang2023tracking}), to enhance the understanding of object and camera movements. We process a video $\mathcal{V}$, created by T2V models using text prompt $\mathcal{T}$, alongside its tracking trajectory $\mathcal{V}_{\mathrm{track}}$ and question-choice pairs $\{Q_i, C_i\}_{i=1}^N$ generated by LLMs. These inputs are then fed into multi-modality LLMs for question answering: $\mathrm{{\hat{A}_i}} = \mathrm{VQA}(\mathcal{V}, \mathcal{V}_{\mathrm{track}}, Q_i, C_i)$.

We define the Text-to-Video (T2V) alignment score \texttt{T2VScore-A} as the accuracy of the video question answering process:
\begin{equation}
\texttt{T2VScore-A}(\mathcal{T}, \mathcal{V}) = \frac{1}{N}\sum^{N}_{i=1}\mathbbm{1}[\hat{A}i=A_i].
\end{equation}
The \texttt{T2VScore-A} ranges from $0$ to $1$, with higher values indicating better alignment between text $\mathcal{T}$ and video $\mathcal{V}$.

\subsection{Video Quality}\label{sec:video_quality}

In this section, we discuss the proposed video quality metric in the T2V Score. Our core principle is simple: it should be able to \textbf{\textit{keep effective}} to evaluate videos from \textit{unseen} generation models that come up after we propose this score. Under this principle, the proposed metric aims to achieve two important goals: \textbf{(G1)} It can more accurately assess the quality of generated videos without seeing any of them (\textbf{zero-shot}); \textbf{(G2)} while adapted to videos generated on known models, it can significantly improve generalized performance on \textbf{unknown models}. Both aims inspire us to drastically improve the generalization ability of the metric, via a combo of Mix-of-Limited-Expert Structure (Sec.~\ref{sec:vqa_moe}), Progressive Optimization Strategy (Sec.~\ref{sec:vqa_opt}), and List-wise Learning Objectives (Sec.~\ref{sec:vqa_obj}), elaborated as follows.
 
\subsubsection{Mix-of-Limited-Expert Structure}
\label{sec:vqa_moe}

Given the \textit{hard-to-explain} nature of quality assessment~\cite{zhang2018unreasonable}, current VQA methods that only learn from human opinions in video quality databases will more or less come with their own biases, leading to poor generalization ability~\cite{mdtvsfa}. Considering our goals, inspired by existing practices~\cite{wu2023dover,wu2023explainable,zhang2022bvqa}, we select two evaluators with different biases as \textit{limited experts}, and fuse their judgments to improve generalization capacity of the final prediction. Primarily, we include a \textbf{\textit{technical expert}} (Fig.~\ref{fig:vqstructure}(a)), aiming at capturing distortion-level quality. This branch adopts the structure of FAST-VQA~\cite{wu2022fastvqa}, which is pre-trained from the largest VQA database, LSVQ~\cite{pvq}, and further fine-tuned on the MaxWell~\cite{wu2023explainable} database that contains a wide range of \textit{spatial} and \textit{temporal} distortions. While the technical branch can already cover scenarios related to naturally-captured videos, generated videos are more likely to include \textit{semantic degradations}, \textit{i.e.} failing to generate correct structures or components of an object. Thus, we include an additional text-prompted \textbf{\textit{semantic expert}} (Fig.~\ref{fig:vqstructure}(b)). It is based on MetaCLIP~\cite{metaclip}, and calculated via a confidence score on the binary classification between the positive prompt \textit{good, high quality} and negative prompt \textit{poor, low quality}. We also add an additional adapter~\cite{clipadapter} to better suit the CLIP-based evaluator in the domain of video quality assessment.

Denote the technical score for the video $\mathcal{V}$ as $\texttt{Q}_\texttt{tech}(\mathcal{V})$, the text-prompted semantic score as $\texttt{Q}_\texttt{sem}(\mathcal{V})$, we fuse the two independently-optimized judgements via ITU-standard perceptual-oriented remapping~\cite{itu}, into the $\texttt{T2VScore-Q}$:
\begin{align}
\mathrm{R}(s) &= \frac{1}{1 + e^{-\frac{s - \mu(s)}{\sigma(s)}}} \\
\texttt{T2VScore-Q}(\mathcal{V}) &= \frac{\mathrm{R}(\texttt{Q}_\texttt{tech})+\mathrm{R}(\texttt{Q}_\texttt{sem})}{2} 
\end{align}
The \texttt{T2VScore-Q} ranges from $0$ to $1$, with higher values indicating better visual quality of video $\mathcal{V}$.



\subsubsection{Progressive Optimization Strategy}
\label{sec:vqa_opt}

\begin{figure}[t!]
    \centering
    \includegraphics[width=0.88\linewidth]{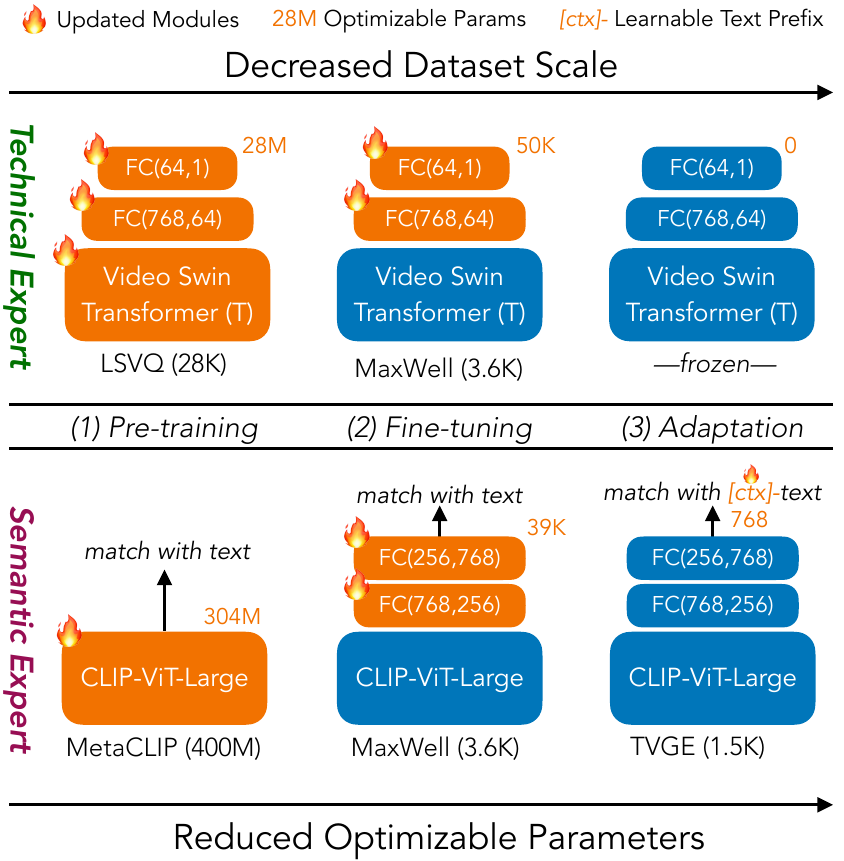}
    \caption{\textbf{Optimization Strategy of the Video Quality Metric,} via gradually decreased scales of training datasets, and correspondingly progressively reduced optimizable parameters.}
    \label{fig:vqoptimize}
    \vspace{-1em}
\end{figure}

After introducing the structure, we discuss the optimization strategy for the \texttt{T2VScore-Q} (Fig.~\ref{fig:vqoptimize}). In general, the training is conducted in three stages: \textit{pre-training}, \textit{fine-tuning}, and \textit{adaptation}. The stages come with gradually smaller datasets, coped with \textbf{progressively reduced optimizable parameters}. For $\texttt{Q}_\texttt{tech}$, the optimization strategies for each stage are listed as follows: (1) \textit{end-to-end} pre-training with large-scale LSVQ$_\text{train}$ dataset (\textit{28K} videos); (2) \textit{multi-layer} fine-tuning with medium-scale MaxWell$_\text{train}$ dataset (\textit{3.6K} videos); (3) Given that specific distortions on generated videos (See Fig.~\ref{fig:domaingap}) are usually associated with semantics, to avoid over-fitting, the technical expert is kept \textit{frozen} during the adaptation stage.
For $\texttt{Q}_\texttt{sem}$, (1) we directly adopt official weights from MetaCLIP~\cite{metaclip} as pre-training; (2) for the fine-tuning stage, we train a lightweight adapter~\cite{clipadapter} on MaxWell$_\text{train}$; (3) for adaptation, we train an additional prefix token~\cite{coop,clipiqa,bvqiplus} to robustly adapt it to the domain of generated videos. 

\subsubsection{List-wise Learning Objectives}
\label{sec:vqa_obj}

Plenty of existing studies~\cite{mdtvsfa,wu2022fastervqa,zhang2022bvqa,wu2022fastvqa} have pointed out that compared with independent scores, the rank relations among different scores are more reliable and generalizable, especially for small-scale VQA datasets~\cite{vqc}. Given these insights, we decide to adopt the \textit{list-wise} learning objectives~\cite{norminnorm} combined by rank loss ($\mathcal{L}_{rank}$) and linear loss ($\mathcal{L}_{linear}$) as our training objective for both limited experts:
\begin{equation}
\mathcal{L}_{rank} = \sum_{i,j} \max((s_{pred}^i - s_{pred}^j)~\mathrm{sgn}~(s_{gt}^j - s_{gt}^i), 0)
\end{equation}
\begin{equation}
\mathcal{L}_{lin}  = (1 - \frac{< s_{pred} - \overline{s_{pred}} ,  s_{gt} - \overline{s_{gt}} >}{\Vert s_{pred} - \overline{s_{pred}} \Vert_2\Vert s_{gt} - \overline{s_{gt}} \Vert_2}) / 2
\end{equation}
\begin{equation}
\mathcal{L} = \mathcal{L}_{lin} + \lambda \mathcal{L}_{rank}
\label{eq:loss}
\end{equation}
where $s_{pred}$ and $s_{gt}$ are \textit{lists} of predicted scores and labels in a batch respectively, and $\mathrm{sgn}$ denotes the sign function.

\section{TVGE Dataset}\label{sec:dataset}

\begin{figure}[t!]
    \centering
    \animategraphics[width=\linewidth]{8}{fig/domain_gap/}{00}{14}
    \caption{\textbf{Domain Gap with Natural Videos.} The common distortions in generated videos (\textit{as in TVGE dataset}) are different from those in natural videos~\cite{wu2023explainable}, both \textit{spatially} and \textit{temporally}. \emph{We encourage readers to \textcolor{magenta}{click and play} using Adobe Acrobat.}}
    \label{fig:domaingap}
    \vspace{-1em}
\end{figure}

\paragraph{Motivation.} An inalienable part of our study is to evaluate the reliability and robustness of the proposed metrics on text-conditioned generated videos. To this end, we propose the \textbf{T}ext-to-\textbf{V}ideo \textbf{G}eneration \textbf{E}valuation (\textbf{TVGE}) dataset, collecting rich human opinions on the two perspectives (\textit{alignment \& quality}) studied in the T2V Score. On both perspectives, the \textbf{TVGE} can be considered as \textit{first-of-its-kind}: First, for the alignment perspective, the dataset will be the first dataset providing text alignment scores rated by a large crowd of human subjects; Second, for the quality perspective, while there are plenty of VQA databases on natural contents~\cite{wu2023explainable,pvq,kv1k}, they show notably different distortion patterns (both \textit{spatially} and \textit{temporally}, see Fig.~\ref{fig:domaingap}) from the generated videos, resulting in an non-negligible \textit{domain gap}. The proposed dataset will serve as a validation on the alignment between the proposed T2V Score and human judgments. Furthermore, it can help our quality metric to better adapt to the domain of text-conditioned generated videos. Details of the dataset are as follows.

\vspace{-1em}
\paragraph{Collection of Videos.} In total, 2543 text-guided generated videos are collected for human rating in the \textbf{TVGE} dataset. These videos are generated by \textit{5} popular text-to-video generation models, under a diverse prompt set as defined by EvalCrafter~\cite{liu2023evalcrafter} covering a wide range of scenarios.

\vspace{-1em}
\paragraph{Subjective Studies.} In the {TVGE} dataset, each video is independently annotated by 10 experienced human subjects from both \textit{text alignment} and \textit{video quality} perspectives. Before the annotation, we trained the human subjects\footnote{Training materials are provided in supplementary materials.} and tested their annotation reliability on a subset of TVGE videos. Each video is rated on a five-point-like scale on either perspective, while examples for each scale are provided in the training materials for subjects.

\begin{figure}[t!]
    \centering
    \includegraphics[width=0.99\linewidth]{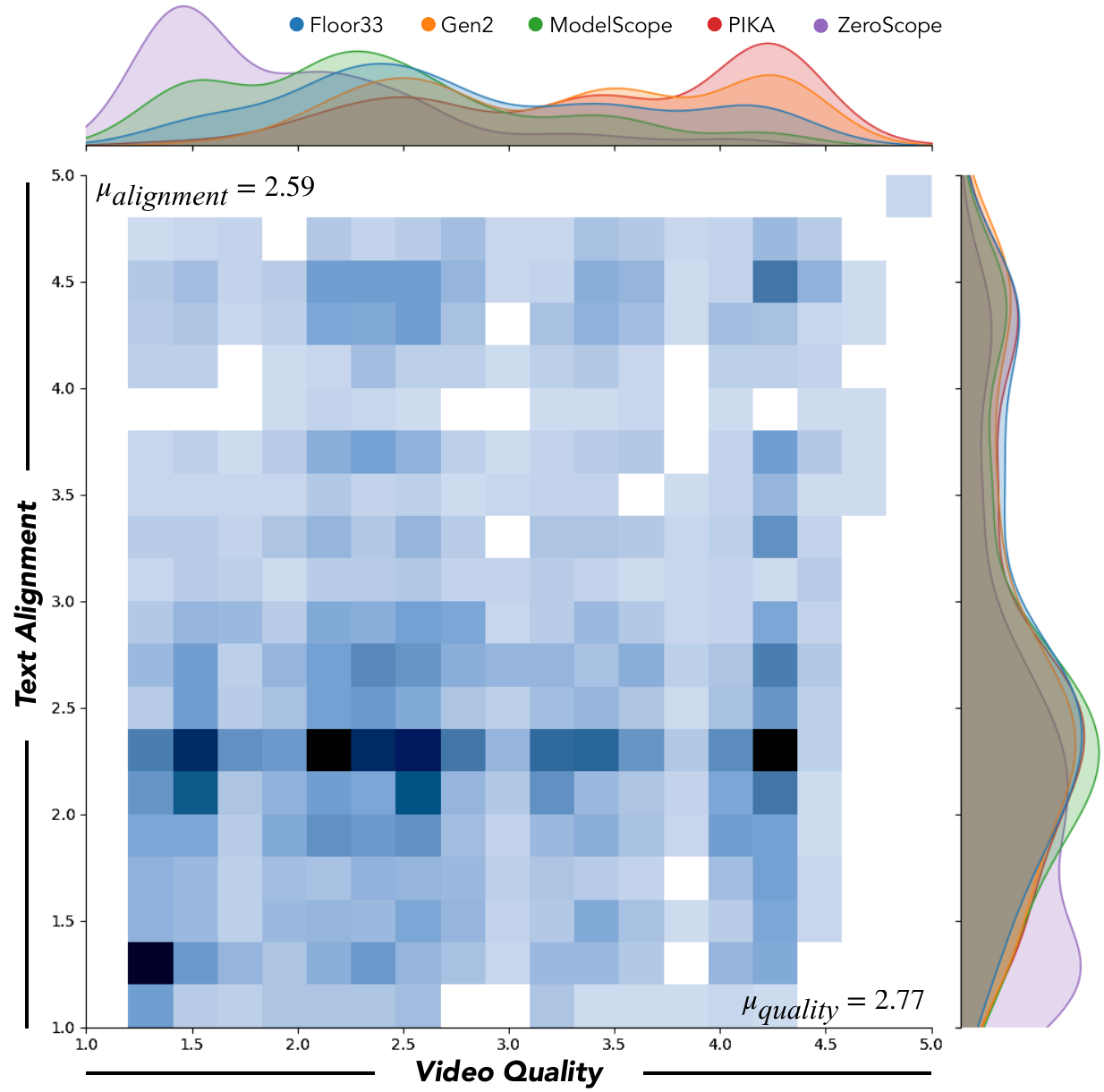}
    \caption{\textbf{Score Distributions in TVGE,} suggesting that current text-to-video generation methods generally face challenges in producing video with either good quality or high alignment with text.  }
    \label{fig:analysis}
    \vspace{-1em}
\end{figure}

\vspace{-1em}
\paragraph{Analysis and Conclusion.} In Fig.~\ref{fig:analysis}, we show the distributions of human annotated \textit{quality} and \textit{alignment} scores in the TVGE dataset. In general, the generated videos receive lower-than-average human ratings ($\mu_\textit{alignment}=2.59$, $\mu_\textit{quality}=2.77$) on both perspectives, suggesting the need to continuously improve these methods to eventually produce plausible videos. Nevertheless, specific models also prove decent proficiency on one single dimension, \textit{e.g.} Pika gets an average score of $3.45$ on \textit{video quality}. Between the two perspectives, we notice a very low correlation (\textit{0.223} Spearman's $\rho$, \textit{0.152} Kendall's $\phi$), proving that the two dimensions are different and should be considered independently. We show more qualitative examples in the supplementary.

\section{Experiments}

\subsection{Text Alignment}

\paragraph{Baselines.}
We compare our \texttt{T2VScore-A} with several standard metrics on text-video alignment, listed as follows:

\begin{itemize}
\item \textit{CLIP Score}~\cite{clip, hessel2022clipscore}: Average text-image similarity in the embedding space of image CLIP models~\cite{clip} over all video frames. 

\item \textit{X-CLIP Score}~\cite{ni2022expanding}: Text-video similarity measured by a video-based CLIP model finetuned on text-video data.

\item \textit{BLIP-BLEU}~\cite{liu2023evalcrafter}: Text-to-text similarity measured by BLEU~\cite{papineni-etal-2002-bleu} score using BLIP-2's image captioning.

\item \textit{mPLUG-BLEU}: Same as \textit{BLIP-BLEU} but using a mPLUG-OWL2 for video captioning.

\end{itemize}

\begin{table}[h]
\centering
\resizebox{\linewidth}{!}{
    \begin{tabular}{cl|cc}
    \toprule
    \multirow{2}{*}{Method} & \multirow{2}{*}{Model} & Spearman's & Kendall's  \\
     &  & $\rho$ & $\tau$ \\
    \hline
    \multirow{4}{*}{Traditional Metric}   
    & CLIP Score & 0.343  &  0.236   \\
    & X-CLIP Score & 0.257  &  0.175   \\
    & BLIP-BLEU & 0.152  & 0.104  \\
    & mPLUG-BLEU & 0.059  & 0.040  \\ 
    \hline
    \hline

    \multirow{6}{*}{\textbf{\texttt{T2VScore-A}}}
    & Otter$^\dag$  & 0.181  & 0.134  \\
    & Video-LLaMA$^\dag$  & 0.288  & 0.206  \\
    & mPLUG-OWL2-V$^\dag$  & \underline{0.394}  & \underline{0.285}  \\
    & InstructBLIP$^*$ & 0.342  & 0.246  \\
    & mPLUG-OWL2-I$^*$  & 0.358  & 0.257  \\
    & \textbf{GPT-4V}$^*$  & \textbf{0.486}  & \textbf{0.360}  \\
    \hline

    \bottomrule
    \end{tabular}
}
\footnotesize{$^\dag$ \textit{via} Video QA; $^*$ \textit{via} Image QA}
\caption{\textbf{Correlation Analysis.} Correlations between objective metrics and human judgment on text-video alignment. Spearman's $\rho$ and Kendall's $\tau$ are used for correlation calculation. The best is \textbf{bold-faced}, and the second-best is \underline{underlined}.}
\label{tab:corr_alignment}
\vspace{-1em}
\end{table}

\vspace{-1em}
\paragraph{Comparison with traditional metrics.} 
We evaluate existing objective metrics using our TVGE dataset and observe a low correlation with human judgment in text-video alignment. This observation aligns with findings in recent research~\cite{liu2023fetv, otani2023toward} that current objective metrics are incompatible with human perception. In particular, video-based CLIP models exhibit even lower correlations than their image-based counterparts in comprehending videos. This discrepancy may be attributed to the X-CLIP score model, which has been fine-tuned exclusively on the Kinetics datasets, a scope insufficient for broad-domain video understanding. Additionally, while BLEU is a widely employed evaluation metric in NLP research, its effectiveness diminishes in text-video alignment tasks. This is due to the inherent challenge of accurate video captioning. Consequently, video-based models such as mPLUG-Owl-2 prove to be less helpful in this context.

\vspace{-1em}
\paragraph{Comparison on MLLMs.}
Our \texttt{T2VScore-A} model is designed to be model-agnostic, which ensures it is compatible with a wide variety of multimodal language learning models (MLLMs). This includes open-source models like Otter~\cite{instructblip}, Video-LLaMA~\cite{damonlpsg2023videollama}, and mPLUG-OWL2-V~\cite{mplugowl2}, as well as proprietary models such as GPT-4V~\cite{GPT4}. In our experiments, we tested \texttt{T2VScore-A} with both image and video-based LLMs. We found that its performance significantly depends on the capabilities of the underlying MLLMs. Open-source image LLMs like InstructBLIP and mPLUG-OWL2-I show decent results in visual question answering. However, their limited temporal understanding makes them less effective compared to the more advanced open-source video LLMs like mPLUG-OWL2-V in video-based question-answering tasks. Despite this, there is still a notable performance disparity between these open-source MLLMs and GPT-4V, with GPT-4V demonstrating superior performance in video question answering. This is evidenced by its higher correlation with human judgment, outperforming other models by a significant margin.

\vspace{-1em}
\paragraph{Effect of auxiliary trajectory.}
We leverage the point trajectory data generated by CoTracker to enhance fine-grained temporal understanding. This approach effectively captures the subtle motion changes of both the object and the camera, which is instrumental in answering questions related to temporal dynamics. As shown in \cref{fig:trajectory}, models that incorporate trajectory data can accurately identify specific camera movements, such as ``panning from right to left" and ``rotating counter-clockwise". In contrast, models without trajectory input struggle to perceive these subtle motion changes. The numerical results in \cref{tab:trajectory} and \cref{tab:vqa_accuracy} further supports our observation.

\subsection{Video Quality}

\paragraph{Baselines.} We compare the \texttt{T2VScore-Q} with several state-of-the-art methods on video quality assessment:
\begin{itemize}
    \item \textit{FAST-VQA}~\cite{wu2022fastvqa}: State-of-the-art technical quality evaluator, with multiple mini-patches (``\textit{fragments}") as inputs.
    \item \textit{DOVER}~\cite{wu2023dover}: State-of-the-art VQA method, consisting of FAST-VQA and an additional aesthetic branch.
    \item \textit{MaxVQA}~\cite{wu2023explainable}: CLIP-based text-prompted VQA method.
\end{itemize}

We also validate the performance of multi-modality foundation models in evaluating generated video quality:
\begin{itemize}
    \item
    \textit{CLIP}~\cite{clip,metaclip}: As CLIP is one of the important bases of the \texttt{T2VScore-Q}, it is important to how original zero-shot CLIP variants work on this task. The original CLIPs are evaluated under the same prompts as the proposed semantic expert: \textit{good, high quality}$\leftrightarrow$\textit{poor, low quality}. 
\end{itemize}

\vspace{-1em}
\paragraph{Settings.} As discussed in Sec.~\ref{sec:video_quality}, we validate the effectiveness of the \texttt{T2VScore-Q} under two settings:
\begin{itemize}
    \item \textbf{(G1):} \textit{zero-shot:} In this setting, no generated videos are seen during model training. Aligning with the settings of off-the-shelf evaluators, it fairly compares between the baseline methods and the proposed \texttt{T2VScore-Q}.
    \item \textbf{(G2):} \textit{adapted, cross-model:} In this setting, we further adapt the \texttt{T2VScore-Q} to a part of the TVGE dataset with videos generated by one \textbf{known} model, and evaluate the accuracy on other 4 \textbf{unknown} generation models. It is a rigorous setting to check the reliability of the proposed metric with future generation models coming.
\end{itemize}

\vspace{-1em}
\paragraph{Comparison on the \emph{zero-shot} setting.} We show the comparison between the \texttt{T2VScore-Q} and baseline methods in Tab.~\ref{tab:zeroshotquality}, under the \textit{zero-shot} setting without training on any generated videos. Firstly, after our fine-tuning (stage 2, on natural VQA dataset), the two experts that make up the \texttt{T2VScore-Q} have notably improved compared with their corresponding baselines; Second, the mixture of the limited experts also resulted in significant performance gain. Both improvements lead to the final more than \textbf{20\%} improvements on all correlation coefficients than existing VQA approaches, demonstrating the superiority of the proposed metric. Nevertheless, without training on any T2V-VQA datasets, all \textit{zero-shot} metrics are still not enough accurate to evaluate the quality of generated videos, bringing the necessity to discuss a robust and effective adaptation approach.

\begin{table}[t]
\centering
\resizebox{\linewidth}{!}{\begin{tabular}{l|ccc}
\toprule
  \multirow{2}{*}{Metric} & Spearman's & Kendall's & Pearson's  \\
 &  $\rho$ & $\phi$ &  $\rho$ \\
\hline
  
FAST-VQA~\cite{wu2022fastvqa} & 0.3518 & 0.2405 & 0.3460 \\
DOVER~\cite{wu2023dover} & 0.3591 & 0.2447 & 0.3587\\
MaxVQA~\cite{wu2023explainable} & 0.4110 & 0.2816 & 0.4002 \\ \hdashline
CLIP-ResNet-50~\cite{clip} & 0.3164 & 0.2162 & 0.3018 \\
CLIP-ViT-Large-14~\cite{metaclip} & 0.3259 & 0.2213 & 0.3140 \\ 
\hline\hline
\textit{the Technical Expert} & 0.4557 & 0.3136 & 0.4426\\
\textit{the Semantic Expert} & 0.4623 & 0.3210 & 0.4353\\ \hdashline
\textbf{\texttt{T2VScore-Q}} (Ours) & \textbf{0.5029} & \textbf{0.3498} & \textbf{0.4945} \\
\hline
\textit{improvements} & +22.3\% & +24.2\% & +23.6\% \\
\bottomrule
\end{tabular}}
\caption{\textbf{\textit{Zero-shot} comparison on Video Quality.} Correlations comparison Spearman's $\rho$, Kendall's $\phi$, and Pearson's $\rho$ are included for correlation calculation.}
\label{tab:zeroshotquality}
\vspace{-1em}
\end{table}

\vspace{-1em}
\paragraph{Cross-model improvements of adaptation.} A common concern on data-driven-generated content quality assessment is that evaluators trained on a specific set of models cannot generalize well on evaluating a novel set of models. Thus, to simulate the real-world application scenario, we abandon the \textit{random} five-fold splits and use rigorous \textit{cross-model} settings during the adaptation stage. As shown in Tab.~\ref{tab:crossmodelquality}, in each setting, we only adopt the \texttt{T2VScore-Q} on videos generated in \textbf{one} among five models in the TVGE dataset and evaluate the changes of accuracy on the rest of videos generated by other 4 models. The table has proven that the proposed prefix-tuning-based adaptation strategy can effectively generalize to \textbf{unseen model sets} with an average of \textbf{11\%} improvements, proving that the \texttt{T2VScore-Q} can be a reliable open-set quality metric for generated videos.

\begin{table}[t]
\centering
\resizebox{\linewidth}{!}{\begin{tabular}{l|ccc}
\toprule
  \multirow{2}{*}{Strategy} & Spearman's & Kendall's & Pearson's  \\
 &  $\rho$ & $\phi$ &  $\rho$ \\
\hline
  
\multicolumn{4}{l}{- Evaluated on other 4 models except \textit{PIKA}} \\ \hdashline
\textit{zero-shot} & 0.4758 & 0.3311 & 0.4643\\ 
Trained on \textit{PIKA}, \textit{cross} & \textbf{0.5467} & \textbf{0.3834} & \textbf{0.5377}\\
\hline
\multicolumn{4}{l}{- Evaluated on other 4 models except \textit{Floor33}} \\ \hdashline
\textit{zero-shot} & 0.5467 & 0.3801 & 0.5363 \\ 
Trained on \textit{Floor33}, \textit{cross} & \textbf{0.5923} & \textbf{0.4192} & \textbf{0.5805} \\
\hline\multicolumn{4}{l}{- Evaluated on other 4 models except \textit{ZeroScope}} \\ \hdashline
\textit{zero-shot} &  0.4148 & 0.2884 &  0.4330 \\ 
Trained on \textit{ZeroScope}, \textit{cross} & \textbf{0.4561} &  \textbf{0.3194} & \textbf{0.4623} \\
\hline\multicolumn{4}{l}{- Evaluated on other 4 models except \textit{ModelScope}} \\ \hdashline
\textit{zero-shot} & 0.4826 & 0.3340 & 0.4835\\ 
Trained on \textit{ModelScope}, \textit{cross} & \textbf{0.5406} & \textbf{0.3785} & \textbf{0.5368}
\\
\hline\multicolumn{4}{l}{- Evaluated on other 4 models except \textit{Gen2}} \\ \hdashline
\textit{zero-shot} & 0.4964 & 0.3472 &0.4920 \\ 
Trained on \textit{Gen2}, \textit{cross} & \textbf{0.5514} & \textbf{0.3895} & \textbf{0.5481}\\
\hline
\textit{average cross-model gain} & +11.2\% & +11.2\% & +10.6\% \\
\bottomrule
\end{tabular}}
\caption{\textbf{\textit{Cross-model} Improvements on Video Quality.} In each setting, we adapt the \textbf{\texttt{T2VScore-Q}} with about 500 videos generated with \textbf{one} of the models, and test its improvements of accuracy on the rest of the videos generated by the other 4 models.}
\label{tab:crossmodelquality}
\end{table}

\vspace{-1em}
\paragraph{Ablation Studies.} We show the ablation experiments in Tab.~\ref{tab:abl}. As is shown in the table, the proposed fine-tuning (stage 2) on both experts improved their single branch accuracy and the overall accuracy of \texttt{T2VScore-Q}, suggesting the effectiveness of the proposed components.

\begin{table}[t]
\centering
\resizebox{\linewidth}{!}{\begin{tabular}{cccc|ccc}
\toprule
\multicolumn{4}{c|}{Components in \texttt{T2VScore-Q}} & Spearman's & Kendall's & Pearson's  \\
\texttt{Q$_\texttt{sem}$} & fine-tune & \texttt{Q$_\texttt{tech}$} & fine-tune
 &  $\rho$ & $\phi$ &  $\rho$ \\
\hline
\hline
\cmark &   &   & & 0.3259 & 0.2213 & 0.3140\\ 
\cmark & \cmark  &   & & 0.4623 & 0.3210 & 0.4353\\ 
   & & \cmark &   & 0.3518 & 0.2405 & 0.3460\\ 
  & & \cmark & \cmark  &  0.4557 & 0.3136 & 0.4426\\  \hdashline
\cmark & \cmark  & \cmark  & & 0.4458 & 0.3074 & 0.4409\\ 
 \cmark & & \cmark & \cmark  &  0.4629 & 0.3197 & 0.4514\\  \hdashline
 \cmark &  \cmark & \cmark & \cmark  & \textbf{0.5029} & \textbf{0.3498} & \textbf{0.4945} \\ \hline
\bottomrule
\end{tabular}}
\caption{\textbf{Ablation Study.} Spearman's $\rho$, Kendall's $\phi$, and Pearson's $\rho$ are included for correlation calculation.}
\label{tab:abl}
\vspace{-1.5em}
\end{table}
\section{Conclusion}

In this paper, to address the shortcomings of existing text-to-video generation metrics, we introduced the Text-to-Video Score (T2VScore), a novel evaluation metric that holistically assesses video generation by considering both the alignment of the video with the input text, and the video quality. Moreover, we present the \textbf{TVGE} dataset to better evaluate the proposed metrics. The experimental results on the TVGE dataset underscore the effectiveness of T2VScore over existing metrics, providing a more comprehensive and reliable means of assessing text-to-video generation. This proposed metric, along with the dataset, paves the way for further research and development in the field aiming at more accurate evaluation methods for video generation methods.

\vspace{-1em}
\paragraph{Limitations and future work.} The \texttt{T2VScore-A} heavily relies on multimodal large language models (MLLMs) to perform accurate Visual Question Answering. However, the current capabilities of MLLMs are not yet sufficient to achieve high accuracy, particularly in evaluating temporal dimensions. We anticipate that as MLLMs become more advanced, our \texttt{T2VScore-A} will also become increasingly stable and reliable.

As new open-source text-to-video models continue to emerge, we will keep track of the latest developments and incorporate their results into our TVGE dataset as part of our future efforts.

{
    \small
    \bibliographystyle{ieeenat_fullname}
    \bibliography{main}
}

\clearpage
\setcounter{page}{1}
\maketitlesupplementary

\section{More Details for Subjective Study}

\begin{figure*}[t]
    \begin{subfigure}{0.48\textwidth}
        \includegraphics[width=\linewidth]{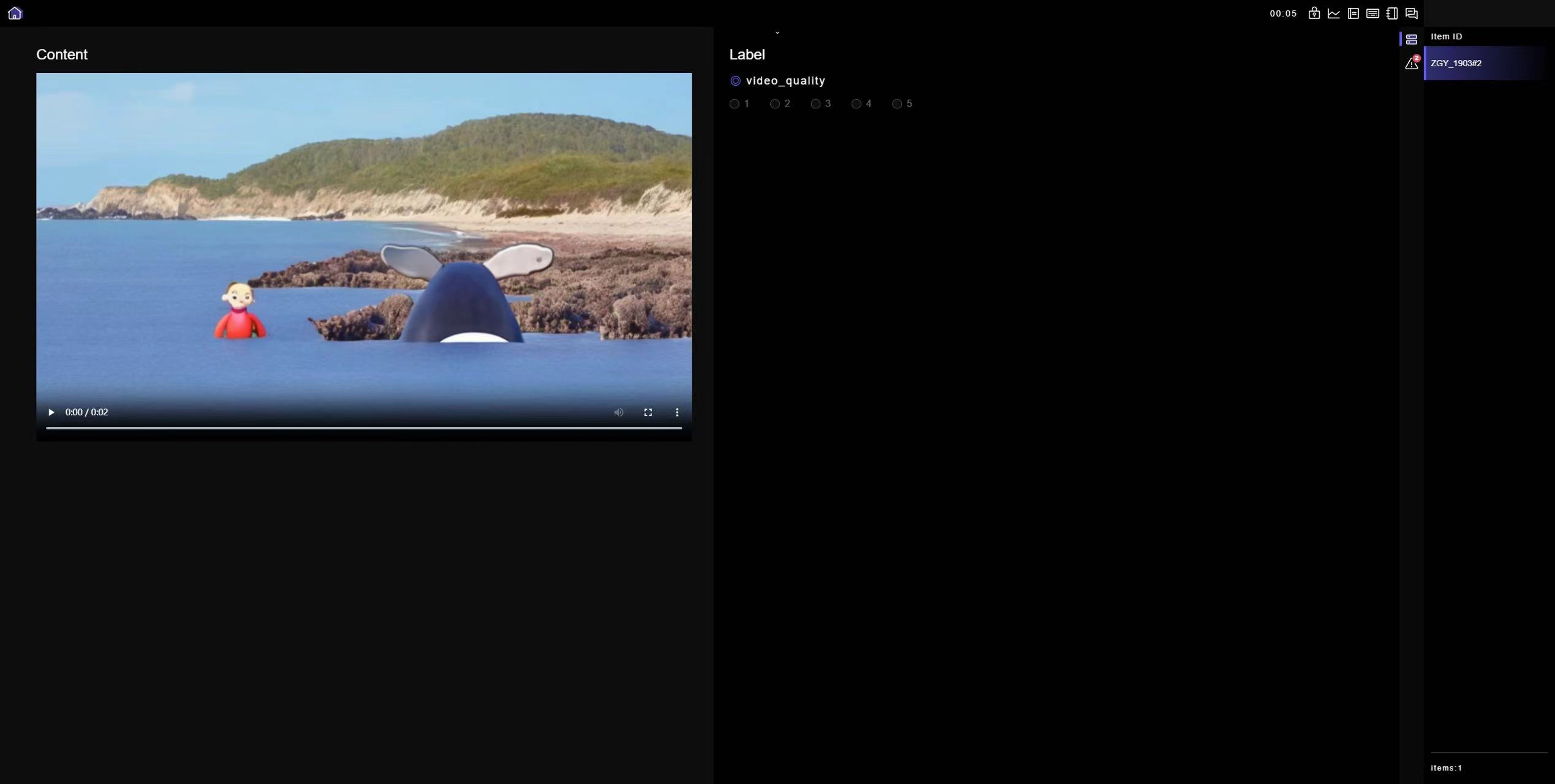}
        \caption{\textbf{Annotation Interface} for \textit{Video Quality}. The video is presented to subjects to be rated a quality score among [1,5].}
    \end{subfigure}
    \hspace{0.02\textwidth}
    \begin{subfigure}{0.48\textwidth}
        \includegraphics[width=\linewidth]{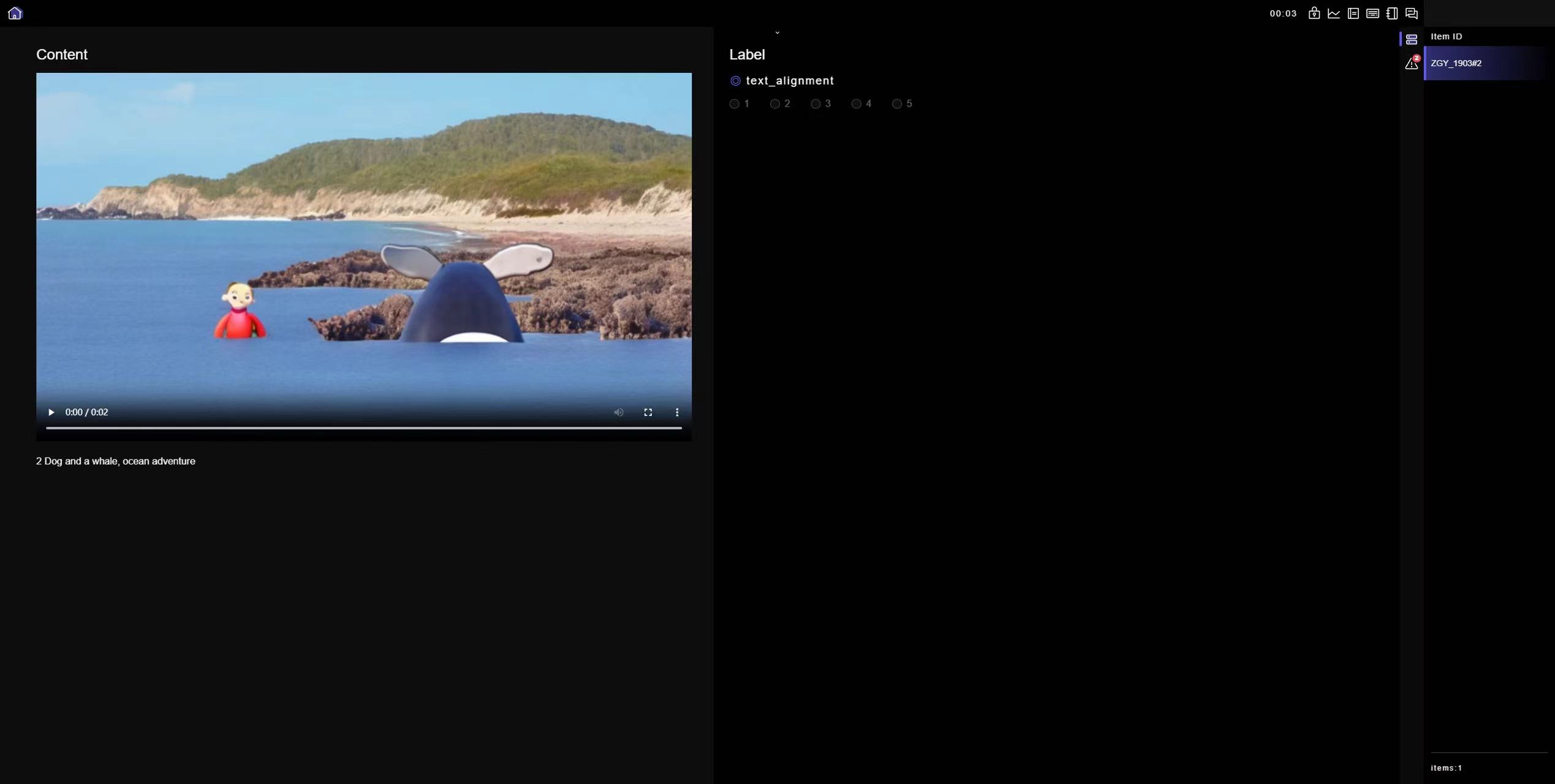}
        \caption{\textbf{Annotation Interface} for \textit{Text Alignment}. The video and its text prompt are presented to subjects to be rated an alignment score among [1,5].}
    \end{subfigure}
    \caption{\textbf{Annotation Interface} for Video Quality (a) and Text Alignment (b).}
    \label{fig:interface}
\end{figure*}

\paragraph{Annotation Interface.} In Fig.~\ref{fig:interface}, we show the annotation interface for the subjective study in the \textbf{TVGE} dataset. The text alignment scores and quality scores are annotated separately to avoid distraction from each other. The input text prompt is shown only for the \textit{alignment} annotation (which is necessary), but not displayed for \textit{quality} annotation, so that the \textit{quality} score can ideally only care about the visual quality of the generated video.

\vspace{-1em}
\paragraph{Training Materials.} Before annotation, we provide clear criteria with abundant examples of 5-point Likert scale 
to train the annotators. For \textbf{text alignment}, we specifically instruct annotators to evaluate the videos based solely on the presence of each element mentioned in the text description, intentionally ignoring video quality. For \textbf{video quality}, we ask the subjects to focus exclusively on technical distortions. We provide five examples for each of the six common distortions: 1) noises; 2) artifacts; 3) blur; 4) unnatural motion; 5) inconsistent structure; and 6) flickering. Samples of the annotated videos can be viewed in \cref{fig:annotation}.

\section{Additional Results}

\paragraph{Effect of Auxiliary Trajectory for \texttt{T2VScore-A}.} As mentioned in \cref{sec:text_alignment}, we utilize the auxiliary point-level trajectory generated by CoTracker~\cite{cotracker} to enhance fine-grained temporal understanding. \cref{fig:trajectory} presents video samples that exhibit temporal nuances, which state-of-the-art multimodal language models (MLLMs) often fail to detect. Using the trajectory as auxiliary information effectively improves the MLLMs' understanding of subtle temporal changes in camera and object motion. For instance, the \textit{snake} in row 5 appears motionless, though the camera is moving. Upon ablating the auxiliary trajectory, we observe a decrease in visual question answering (VQA) accuracy from 0.58 to 0.48, as shown in \cref{tab:vqa_accuracy}. This reduction in VQA accuracy further leads to a diminished alignment with human judgment (see \cref{tab:trajectory}). 

\begin{table}[h]
    \centering
    \resizebox{\linewidth}{!}{\begin{tabular}{l|ccc}
    \toprule
      \multirow{2}{*}{\texttt{T2VScore-A}} & Spearman's & Kendall's & Pearson's  \\
     &  $\rho$ & $\phi$ &  $\rho$ \\
    \hline
      
    GPT-4V \textit{w/o tarjectory} & 0.4454 & 0.3289 & 0.4416 \\
    GPT-4V & \textbf{0.4859} & \textbf{0.3600} & \textbf{0.4882} \\
    \bottomrule
    \end{tabular}}
    \caption{\textbf{Effect of Auxiliary Trajectory.} Spearman's $\rho$, Kendall's $\phi$, and Pearson's $\rho$ are included for correlation calculation.}
    \label{tab:trajectory}
    \vspace{-2em}
\end{table}

\begin{figure}[t]
    \centering
    \animategraphics[width=\linewidth]{8}{fig/trajectory/}{00}{14}
    \caption{\textbf{Quantitative Examples for Auxiliary Trajectory.} Using an auxiliary trajectory effectively enhances multimodal large language models (MLLMs) for fine-grained temporal understanding.
    \emph{Please \textcolor{magenta}{click and play} using Adobe Acrobat.}}
    \label{fig:trajectory}
    \vspace{-1em}
\end{figure}

\begin{table}[t]
    \centering
    \resizebox{\linewidth}{!}{\begin{tabular}{l|ccc}
    \toprule
    Model & Overall & Temporal QA & Spatial QA  \\
    \hline
    
    \textit{random guess} & 0.2369 & 0.2452 & 0.2327 \\
    \hdashline
    Otter$^\dag$ & 0.1460 & 0.1059 & 0.1636 \\
    Video-LLaMA$^\dag$ & 0.4074 & 0.3459 & 0.4435 \\
    mPLUG-OWL2-V$^\dag$ & 0.5305 & 0.4280 & 0.5880 \\
    InstructBLIP$^*$ & 0.5013 & 0.4762 & 0.5127 \\
    mPLUG-OWL2-I$^*$ & 0.5107 & 0.4333 & 0.5600 \\
    \hdashline
    GPT-4V$^*$ \textit{(w/o trajectory)} & 0.4791 & 0.4411 & 0.5589 \\
    GPT-4V$^*$ & \textbf{0.5765} & \textbf{0.5077} & \textbf{0.6308} \\
    \bottomrule
    \end{tabular}}
    \footnotesize{$^\dag$ \textit{via} Video QA; $^*$ \textit{via} Image QA}
    \vspace{-0.5em}
    \caption{\textbf{Accuracy of Visual Question Answering.}}
    \label{tab:vqa_accuracy}
    \vspace{-2em}
\end{table}

\paragraph{Performance of state-of-the-art MLLMs in VQA.} We setup an evaluation set of 500 videos (100 prompts with 5 unique videos per prompt) sampled from our TVGE dataset. Two annotators are tasked with responding to the generated questions, and a third, more experienced annotator is assigned to verify these responses.
We compare the accuracy of visual question answering (VQA) across a range of multimodal large language models (MLLMs), focusing on \textit{spatial} and \textit{temporal} QA. As shown in \cref{tab:vqa_accuracy}, current MLLMs generally demonstrate weak performance in open-domain VQA tasks, with temporal QA faring even worse. Notably, video-based MLLMs are inferior in temporal QA compared to their image-based counterparts. A similar observation is made in SEED-Bench~\cite{seed-bench}, indicating significant room for further improvement in video-based MLLMs. 

\begin{figure*}[t!]
    \centering
    \animategraphics[width=\linewidth]{8}{fig/annotation/}{00}{08}
    \caption{\textbf{Human Annotation.} Generated videos and their human ratings of \textit{text alignment} and \textit{video quality}. The scores are the mean of 10 annotators’ ratings.
    \emph{Please \textcolor{magenta}{click and play} using Adobe Acrobat.}}
    \label{fig:annotation}
    \vspace{2em}
\end{figure*}

\begin{figure*}[t!]
    \centering
    \animategraphics[width=\linewidth]{8}{fig/text_alignment_more/}{00}{08}
    \caption{\textbf{More Examples for \texttt{T2VScore-A}.} We showcase more examples illustrating how \texttt{T2VScore-A} is computed.
    \emph{Please \textcolor{magenta}{click and play} using Adobe Acrobat.}}
    \label{fig:rating}
    \vspace{2em}
\end{figure*}

\begin{figure*}[t]
    \centering
    \includegraphics[width=0.99\linewidth]{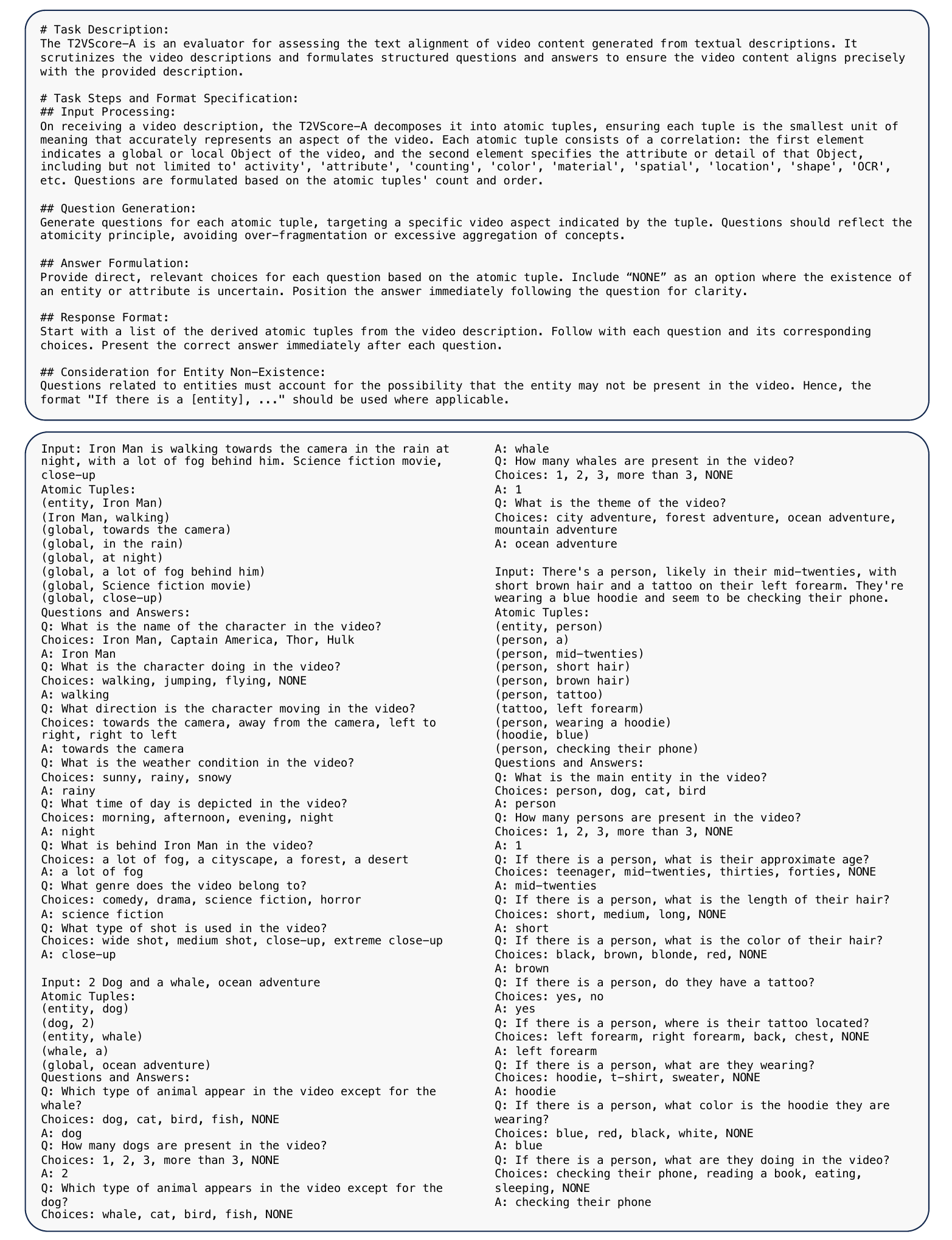}
    \caption{\textbf{Prompt for Question/Answer Generation in \texttt{T2VScore-A}.} \textit{Top:} task instruction; \textit{Bottom:} in-context learning examples.}
    \label{fig:prompt}
\end{figure*}

\end{document}